\documentclass{article}

\usepackage{arxiv}

\usepackage[utf8]{inputenc} 
\usepackage[T1]{fontenc}    
\usepackage{hyperref}       
\usepackage{url}            
\usepackage{booktabs}       
\usepackage{amsfonts}       
\usepackage{nicefrac}       
\usepackage{microtype}      
\usepackage{amsmath}
\usepackage{cleveref}       
\usepackage{lipsum}         
\usepackage{graphicx}
\usepackage{natbib}
\usepackage{doi}
\usepackage{subcaption}
\usepackage{pdflscape}
\usepackage{multirow} 
\usepackage{float} 
\usepackage{graphicx}

\newgeometry{landscape}
\title{MULTISTAGE NON-DETERMINISTIC CLASSIFICATION USING
SECONDARY CONCEPT GRAPHS AND GRAPH
CONVOLUTIONAL NETWORKS FOR HIGH-LEVEL FEATURE
EXTRACTION}

\newif\ifuniqueAffiliation

\ifuniqueAffiliation 
...

\else
\usepackage{authblk}

\newbox{\orcid}\sbox{\orcid}{\includegraphics[scale=0.06 , keepaspectratio]{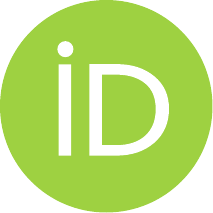}} 

\author[1,2]{\href{https://orcid.org/0000-0002-6650-3538}{\usebox{\orcid}\hspace{1mm}Masoud Kargar}\thanks{\texttt{kargar@iaut.ac.ir}}}
\author[1,2]{\href{https://orcid.org/0009-0006-6325-1538}{\usebox{\orcid}
\hspace{1mm}Nasim Jelodari}\thanks{\texttt{nasimjelodari@iaut.ac.ir}}}
\author[1,2]{\href{https://orcid.org/0009-0008-1797-3606}{\usebox{\orcid}\hspace{1mm}Alireza Assadzadeh}\thanks{\texttt{alireza.asadzadeh@iaut.ac.ir}}}

\affil[1]{Department of Computer Engineering, Islamic Azad University, Tabriz Branch, Iran}
\affil[2]{Robotics and Soft Technologies Research Center, Islamic Azad University, Tabriz Branch, Iran}

\fi

\renewcommand{\shorttitle}{\textit{arXiv} }

\hypersetup{
pdftitle={MULTISTAGE NON-DETERMINISTIC CLASSIFICATION USING
SECONDARY CONCEPT GRAPHS AND GRAPH
CONVOLUTIONAL NETWORKS FOR HIGH-LEVEL FEATURE
EXTRACTION},
pdfsubject={q-bio.NC, q-bio.QM},
pdfauthor={David S.~Hippocampus, Elias D.~Striatum},
pdfkeywords={First keyword, Second keyword, More},
}

\begin{document}
\maketitle

\begin{abstract}
	Graphs, comprising nodes and edges, visually depict relationships
and structures, posing challenges in extracting high-level features due to their
intricate connections Multiple connections introduce complexities in discovering
patterns, where node weights may affect some features more than others.
In domains with diverse topics, graph representations illustrate interrelations
among features.Pattern discovery within graphs is recognized as NPhard.
Graph Convolutional Networks(GCNs)are a prominent deep learning approach
for acquiring meaningful representations by leveraging node connectivity
and characteristics.Despite achievements, predicting and assigning 9 deterministic
classes often involve errors.To address this challenge, we present
a multi-stage non-deterministic classification method based on a secondary
conceptual graph and graph convolutional networks, which includes distinct
steps:1)leveraging GCN for the extraction and generation of 12 high-level features;
2) employing incomplete, non-deterministic models for feature extraction,
conducted before reaching a definitive prediction;3)formulating definitive
forecasts grounded in conceptual (logical)graphs. The empirical findings indicate
that our proposed approach outperforms contemporary methods in classification
tasks.Across three datasets—Cora, Citeseer, and PubMed—the achieved
accuracies are 96\%, 93\%, and 95\%, respectively.
Code is available at https://github.com/MasoudKargar
\end{abstract}


\section{Introduction}
In recent decades, with the significant expansion of networks and the vast amount of data generated by these networks, the demand for graphs to effectively display these data has increased dramatically \citep{goyal2018graph} . As an essential data structure, a graph includes a set of nodes and edges. This data structure effectively describes and visualizes the relationships between nodes and edges \citep{waikhom2021graph, jiang2020co}. Graphs have many applications in various fields, including social networks, sensor networks, and document networks \citep{pan2019learning}. Feature-based graphs are another type of graphs where the nodes and edges have information or, in other words, features. Due to the extensive and complex interactions of information within networks, graph analysis is of particular importance.\\
Classification is considered a challenge in feature-based graphs due to their unique features. Node and graph classification are two particular challenges in graph learning that have been recently noticed. In machine learning, classification is a common task that aims to classify nodes in a graph and predict their corresponding class labels. Furthermore, graph classification involves predicting class labels assigned to entire graphs \citep{li2019semi, ramanath2018towards}.\\
Due to the high computational cost and space, analyzing these graphs, obtaining the structural relationship between the nodes, and exploiting the node content information are essential challenges. Recent research endeavors have employed graph embedding techniques to tackle this challenge \citep{cai2018comprehensive, wang2022deep}. The primary objective of graph embedding techniques is to depict graph data in a lower-dimensional vector space, ensuring the preservation of its inherent structural characteristics. These embeddings find versatile applications across various tasks, including node classification, link prediction, and more. They function as input features for classification models, enabling the prediction of class labels for the graph. Besides graph embedding, alternative types of embeddings, such as node, edge, and infrastructure embeddings, exist \citep{xu2021understanding}. Deep learning methods, notably graph neural networks (GNN) and graph convolutional networks (GCN), have emerged as prominent techniques for analyzing and interpreting graphs \citep{xiao2022graph,zeng2019accurate}. The core concept behind GCNs involves leveraging a graph structure to recognize the neighboring nodes of each node and acquire the node embedding by recursively aggregating the embeddings of its neighbors \citep{yu2021knowledge}.\\
One of the important challenges in this field is that the samples in the used data set may be members of several classes instead of being members of one class. In this case, making predictions and assigning to a definite class is usually associated with many errors.\\
We propose a multistage non-deterministic classification based on a secondary conceptual graph and graph convolutional networks for node classification. The proposed architecture consists of several steps: 1) Using GCN to extract and generate high-level features from code and embedded inputs. 2) Using incomplete (non-deterministic) models for non-deterministic forecasting. 3) creating a secondary conceptual graph based on non-deterministic prediction, and 4) performing classification based on conceptual (logical) graph. According to Figure 1, in our proposed method, between the stages of high-level feature extraction using GCN and final classification, we first perform a non-deterministic prediction and then create a conceptual layer to create a conceptual graph. By creating and applying these steps in multi-class datasets, classification accuracy increases. The main contributions of this paper are:\\
• We presented a multi-stage non-deterministic classification framework based on secondary conceptual graph and graph convolution networks, which is a combination of several stages.\\
• We use two GCN layers and then make a non-deterministic prediction by the conceptual layers we created before classification to improve the classification performance.\\
• Experimental results show that our method works better than node classification methods.\\

\begin{figure}[h]
	\centering
	\includegraphics[height=0.2\textheight , keepaspectratio]{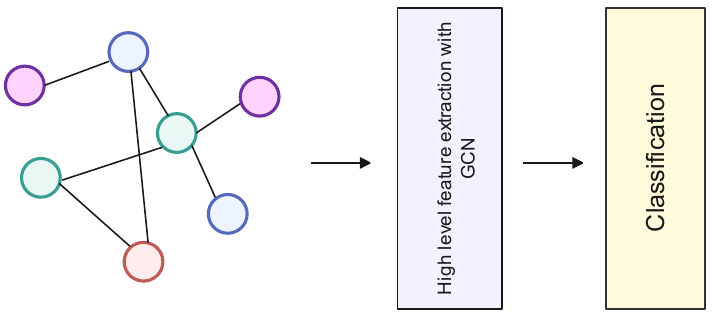}
	\caption{Classification based on GCN}
	\label{fig:fig1}
\end{figure}

RQ1: Can models based on incomplete models be effective in improving forecasting?\\
RQ2: Can intermediate logic graphs affect the accuracy of the final prediction?\\
RQ3: Can non-deterministic stage prediction be done with incomplete hidden layers?\\
This research is organized into five chapters: Section 2 introduces the related work, and Section 3 presents the proposed methodology. Section 4 reviews the experimental data and research results; section 5 describes the research conclusions and discusses future works.

\section{Previous Work}
Classification has emerged as a prevalent task in machine learning, focusing on classification nodes or subgraphs within a graph into discrete classes or labels. This is particularly valu-able when there is a need to comprehend the connections and patterns among elements in a graph. Recently, graph neural networks have demonstrated substantial progress in node classification, graph classification, and link prediction tasks.
\subsection{Node classification}
We will provide a brief overview of the three categories into which graph neural networks are classified based on the mechanism they use for node classification.
FSGNN, a simple two-layer model for graph neural networks (GNN), was presented by Liu et al. FSGNN incorporates a ”soft selection” mechanism that effectively learns the importance of features during the training process. The proposed approach performs better than the state-of-the-art method (SOTA) and achieves more than 51.1\% accuracy in node classification in the given class  \cite{maurya2022simplifying}. Yang Hu et al. introduced a simple graph learning framework called graph-MLP, which relies on MLP (Multilayer Perceptron) with-out message-passing modules. Notably, this framework achieves graph node classification for the first time without using traditional message-passing methods. What distinguishes
graph-MLP is its ability to maintain stable performance even when adjacency information is compromised or lost during the inference process \cite{hu2021graph}.
Yao et al. proposed a new framework for joint training that incorporates pairwise constraints between nodes and uses the AdaDW strategy to enhance the training of classification models in homogeneous graphs. However, it should be noted that this method does not apply to heterogeneous graphs such as HLN, making it ineffective in such scenarios \citep{wu2021enhancing}.To address the limited information acquisition caused by the shallow structure of existing classification methods, a hierarchical graph attention network (HGAT) is introduced for semi-supervised classification. HGAT increases the accuracy by combining the attention mechanism in the input and prediction layers and assigning weights to different
nodes. The simulation results show the improved performance of the proposed method on simplified datasets such as Cora, Citeseer, PubMed, and NELL. However, the limitation of this approach is its lack of application exclusively for directed graphs \citep{li2020hierarchical}.
Tang et al. introduce a new GNN-INCM model, including ECO and GRO modules for node embedding learning. This paper also uses HSKDM to train multiple GNN-INCM models simultaneously. GNN-INCM shows improved performance on the base dataset, and using HSKDM further increases the classification performance \cite{huang2022graph}.
Yu Wang et al. proposed a tree decomposition method to separate neighborhood features into distinct layers. They introduce a tree-decomposed graph neural network (TDGNN),which includes TDGNN-s and TDGNN-w variants for node classification. This method provides a flexible layer configuration for complex networks and reduces the feature smoothing problem \citep{wang2021tree}. Tang et al. introduce a new analytical framework called WCE-GNN that explores the relationship between graph neural networks (GNN) and hypergraph neural networks (HyperGNN) in the context of node classification. The framework enables the direct application of GNNs for hypergraph node classification. The results indicate that WCE-GNN has lower spatial and temporal complexity than modern HyperGNN methods \citep{tang2024hypergraph}.
Wang et al. have introduced a novel Graph Convolutional Network (GCL) encoder referred to as low graph adversarial learning (LR GCL) .The proposed methodology engages in semi-supervised or transitional node classification through a two-stage process, wherein the graph data undergoes corruption by noise in the feature labels of each node \citep{wang2024low}. Sejan et al. have introduced a model based on Graph Convolutional Networks (GCN), utilizing weight, adjacency, and identity matrices to enhance node classification performance, specifically regarding classification accuracy \citep{sejan2023graph}.

A novel Community GCN method is introduced for community detection in social networks through node classification utilizing Graph Convolutional Networks (GCN). This approach adeptly leverages the benefits of transmitting messages via feedforward networks (FFN)and traversing connections via GCN. Furthermore, it incorporates spatial proximity matrices, weighted average feature extraction, and GCNN-based node classification, effectively contributing to identifying communities within social networks \citep{bhattacharya2023communitygcn}.
\subsection{Graph classification}
Lara et al. introduced a simple algorithm that uses Laplacian graph spectral decomposition to achieve graph classification and provide an initial reference score for a given dataset \citep{de2018simple}. The GAM framework, a new RNN-based approach for graph classification with attention, is presented. This framework uses external memory to
integrate information from different parts of the graph and takes care of processing parts of the graph \citep{lee2018graph}.
Similarly, Wang et al. propose an innovative edge feature graph attention network (EGAT)that combines edge and node data into a graph attention mechanism. This study is the first to consider edge features in the graph attention mechanism. EGAT is a versatile framework that integrates custom messaging functions to address specific domains. The simulation
results in this paper show that the proposed method outperforms previous approaches in terms of accuracy when evaluated on Cora, Citeseer, PubMed, and AMLSim datasets \citep{wang2021egat}. To address the issue of over-smoothing resulting from an augmented number of convolutional layers in graph classification tasks, the authors advocate for implementing
a Multi-Level Coarsening-based Graph Convolutional Network (MLC-GCN). This framework introduces an adaptive module for structural coarsening to produce coarse graphs,subsequently employed in constructing the convolutional network. MLC-GCN stands out by acquiring the ability to represent graphs at various hierarchical levels while retaining
local and global information \citep{xie2020graph}.
\subsection{Edge classification}
Jiang et al. introduced CensNet, a convolution with node-edge switching graph neural network designed for learning tasks involving graph-structured data with node and edge features. CensNet is a versatile graph embedding framework that embeds nodes and edges in a hidden feature space. The proposed method performs strongly in various graph learning tasks such as semi-supervised node classification, graph classification, and unsupervised link prediction, outperforming existing approaches in various benchmarks and establishing itself as a state-of-the-art solution \citep{jiang2020co}.
A new unsupervised approach called AttE2vec is introduced, aiming to learn a low-dimensional vector representation for attributed edges. AttE2vec directly obtains feature information from edges and nodes, even with significant distances, to infer current and new node embeddings. The proposed method shows good edge classification and clustering results in Cora,Citeseer, and PubMed datasets \cite{bielak2022attre2vec}. Zhong et al. present a dynamic graph representation learning technique based on a time transformer graph (TGT). TGT uses a sequential interactive switching network to learn information from neighboring nodes at one-step and two-step intervals, ensuring accurate node representation. This article intro-
duces three aggregation modules and one propagation module, which increases the accuracy of dynamic graph display while reducing execution time. This method works well in link prediction and edge classification \cite{zhong2023dynamic}.

\begin{landscape}
\begin{table}[h]
  \caption{Review and classification of articles in the second part}
  \label{tbl:wasss_citeseer}
  \resizebox{\linewidth}{!}{%
    \begin{tabular}{|c|c|p{0.8cm}|p{3.5cm}|p{3.5cm}|p{3.5cm}|p{4.5cm}|p{4.5cm}|}
      \hline
      \multirow{39}{*}{\rotatebox{90}{Node Classification}} & \textbf{Ref} & \textbf{Year} & \textbf{Title} & \textbf{Authors} & \textbf{Datasets} & \textbf{Advantages} & \textbf{Disadvantages} \\
      \cline{2-8}
      & \citep{maurya2022simplifying} & 2022 & Simplifying approach to node classification in Graph Neural Networks & Sunil Kumar Maurya, Xin Liu, Tsuyoshi Murata & Cora, Citeseer, PubMed, Chameleon, Wisconsin, Texas, Cornell, Squirrel, Acto & Increase accuracy, avoiding over-smoothing phenomenon & Emphasizes the potential downside of utilizing less informative features derived from aggregation processes, which may detrimentally affect the effectiveness of node classification \\
      \cline{2-8}
      & \citep{hu2021graph} & 2021 & Graph-MLP: Node Classification without Message Passing in Graph & Yang Hu, Haoxuan You, Zhe can Wang, Zhicheng Wang, Erjin Zhou, Yue Gao & Cora, Citeseer, PubMed & Effectively perform graph node classification tasks without explicit message passing modules, better performance in graph node classification task & \\
     \cline{2-8}
      & \citep{wu2021enhancing} & 2021 & Enhancing Graph Neural Networks via auxiliary training for semi-supervised node classification & Yao Wu, Yu Song, Hong Huang, Fanghua Ye, Xing Xie, Hai Jin & Cora, Citeseer, PubMed & Higher classification accuracy, low computational cost & \\
      \cline{2-8}
      & \citep{li2020hierarchical} & 2020 & Hierarchical graph attention networks for semi-supervised node classification & Yixiong Feng, Kangjie Li, Yicong Gao, Jian Qiu & Cora, Citeseer, PubMed, Simplified NEL & No need for expensive matrix operations & Not directly applied to a directed graph, not considering large datasets \\
      \cline{2-8}
      & \citep{huang2022graph} & 2022 & A graph neural network-based node classification model on class-imbalanced graph data & Zhenhua Huang, Yinhao Tang, Yunwen Chen & Cora, Citeseer, PubMed & Improve the overall classification performance, optimize two cooperative modules, ECO and GRO & Not considering class-imbalanced graph data \\
      \cline{2-8}
      & \citep{wang2021tree} & 2021 & Tree Decomposed Graph Neural Network & Yu Wang, Tyler Derr & Cora, Citeseer, PubMed & Solving the problem of feature smoothing between different layers and incorporating the multi-hop dependency & Not fully addressing the challenge of generalizing to unseen graphs \\
      \cline{2-8}
      & \citep{tang2024hypergraph} & 2024 & Hypergraph Node Classification With Graph Neural Networks & Bohan Tang, Zexi Liu, Keyue Jiang, Siheng Chen, Xiaowen Dong & Cora, CiteSeer, PubMed, Cora-CA, DBLP-CA, Congress, Senate, Walmart, House & Can capture more complex relationships between nodes compared to traditional graph representations & Does not explicitly consider the hierarchical structure of the graph \\
      \hline
    \end{tabular}%
  }
\end{table}
\end{landscape}

\begin{landscape}
\begin{table}[htbp]
  \ContinuedFloat
  \caption{Review and classification of articles in the second part (continued)}
  \resizebox{\linewidth}{!}{%
    \begin{tabular}{|c|c|p{0.8cm}|p{3.5cm}|p{3.5cm}|p{3.5cm}|p{4.5cm}|p{4.5cm}|}
      \hline
      \multirow{20}{*}{\rotatebox{90}{Node Classification}}
      & \citep{wang2024low} 
      & 2024 
      & Low-Rank Graph Contrastive Learning for Node Classification
      & Yancheng Wang, Yingzhen Yang
      & Cora, CiteSeer, PubMed, Coauthor CS, ogbn-arxiv, Wiki-CS, Amazon-Computers, Amazon-Photos
      & Producing representations with high generalizability, transferability, and robustness, even without sophisticated GNN architectures
      & Not fully address the challenge of noise in real-world graph data, which can significantly degrade the performance of GNN \\
      \cline{2-8}
      & \citep{sejan2023graph} 
      & 2023 
      & Graph Convolutional Network Design for Node Classification Accuracy Improvement
      & Mohammad Abrar Shakil Sejan, Md Habibur Rahman, Md Abdul Aziz, Jung-In Baik, Young-Hwan You, Hyoung-Kyu Song
      & Cora, Citeseer, PubMed
      & Achieves comparable results in the node classification task
      & Inability to accurately predict node classifications as the number of classes grows \\
      \cline{2-8}
      & \citep{bhattacharya2023communitygcn} 
      & 2023 
      & Community GCN: community detection using node classification with graph convolution network
      & Riju Bhattacharya, Naresh Kumar Nagwani, Sarsij Tripathi
      & CiteSeer, Cora, ACM, Karate, FB, IMDB
      & Efficiently and accurately detect communities in large-scale networks
      & Using semi-supervised learning in this model may limit its applicability to specific scenarios \\
      \hline
      \multirow{16}{*}{\rotatebox{90}{Edge Classification}}
      & \citep{jiang2020co} 
      & 2020 
      & Co-embedding of Nodes and Edges with Graph Neural Network learning
      & Xiaodong Jiang, Ronghang Zhu, Pengsheng Ji, Sheng Li
      & Cora, Citeseer, PubMed
      & Enhance the node embeddings and improve the edge embeddings
      & Not considering more graph layers, massive graphs, and dynamic neural networks \\
      \cline{2-8}
      & \citep{bielak2022attre2vec} 
      & 2022 
      & AttrE2vec: Unsupervised attributed edge representation learning
      & Piotr Bielak, Tomasz Kajdanowicz, Nitesh V. Chawla
      & Cora, Citeseer, PubMed
      & Building more powerful edge vector representations, increase AUC and accuracy
      & Not fully address the challenge of capturing complex relational information and long-range dependencies between nodes in the graph \\
      \cline{2-8}
      & \citep{zhong2023dynamic} 
      & 2023 
      & A dynamic graph representation learning based on the temporal graph transformer
      & Ying Zhong, Chenze Huang
      & UCI, MovieLens-10M, Mathoverflow
      & Reduces time consumption, improves performance.
      & The limitation of modeling dynamic graph with fixed nodes and not including structural and temporal information\\
      \hline
      \multirow{22}{*}{\rotatebox{90}{\centering Graph Classification}}
      & \citep{chen2023egc2} 
      & 2023 
      & EGC2: Enhanced graph classification with easy graph compression
      & Jinyin Chen, Haiyang Xiong, Haibin Zheng, Dunjie Zhang, Jian Zhang, Mingwei Jia, Yi Liu
      & PTC, PROTEINS, DD, NCI1, NCI109, IMDB-BINARY, REDDIT-BINARY, OGBL-MOLHIV
      & Efficient and robust graph classification performance, excellent transferability, not causing a noticeable increase in complexity
      & Not fully addressing the challenge of high complexity and lack of transferability faced by many existing graph classification models \\
      \cline{2-8}
      & \citep{de2018simple} 
      & 2018 
      & A Simple Baseline Algorithm for Graph Classification
      & Nathan de Lara, Edouard Pineau
      & MT, PTC, EZ, PF, DD, NCI1
      & Simplicity, efficiency, and competitive performance in latent graph classification tasks
      &  \\
      \cline{2-8}
      & \citep{lee2018graph} 
      & 2018 
      & Graph Classification using Structural Attention
      & John Boaz Lee, Ryan Rossi, Xiangnan Kong
      & HIV, NCI-1, NCI-33, NCI-83, NCI-123
      & The attention mechanism is easily parallelizable
      & Limited to a part of the graph \\
      \cline{2-8}
      & \citep{wang2021egat} 
      & 2021
      & EGAT: Edge-Featured Graph Attention Network
      & Ziming Wang, Jun Chen, Haopeng Chen
      & Cora, Citeseer, PubMed
      & Achieves on both node-sensitive and edge-sensitive datasets
      & Not considering edge features \\
      \cline{2-8}
      & \citep{xie2020graph} 
      & 2020 
      & Graph convolutional networks with multi-level coarsening for graph classification. Knowledge-Based Systems
      & Yu Xie, Chuanyu Yao, Maoguo Gong, Cheng Chen, A.K. Qin
      & D\&D ENZYMES,IMDB-B,IMDB-M, MUTAG, PROTEINS,REDDIT-B,REDDIT-M
      & 
      & Failing to tackle the issue of over-smoothing, which hinders the model's capacity to capture hierarchical information and global patterns within graphs \\
      \hline
    \end{tabular}%
  }
\end{table}
\end{landscape}

\section{Proposed method}\label{se:intro}

An attributed graph is denoted by H = (V, E), where V is a set of nodes, and E is a set of edges. In H, each node v is associated with a feature vector xv. In the feature vector xv, m shows the number of node features. Furthermore, the structure of the graph is represented by the adjacency matrix A, where Aij = 1 if there is an edge between nodes vi and vj , and
Aij = 0 otherwise. The feature matrix X is an n×m matrix, where n represents the number of samples and m represents the number of features. In the attributed graphs, there is a discussion called neighborhood. In other words, the neighborhood represents the common feature among the nodes. As shown in Figure 1, each node has a specific color, each color
represents a group; for example, the orange color refers to one group, and the green color to another group.
\begin{figure}[h]
	\centering
	 \includegraphics[height=0.2\textheight ]{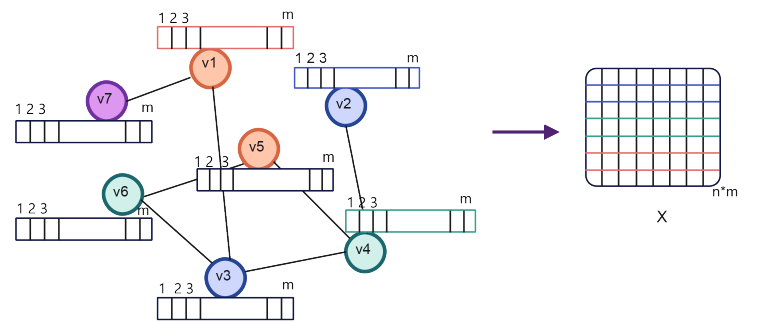}
    \caption{Attributed Graph with a Feature matrix}
	\label{fig:fig1}
\end{figure}
\subsection{Architecture}\label{se:intro}
Figure 3 shows the general framework proposed in this study. Our approach consists of several steps.The first step is graph input, which includes nodes and edges that show the data structure and their relationships. The second step is extracting high-level features from nodes by GCN. The third step is the non-deterministic prediction step, which uses incomplete and non-deterministic models to extract intermediate features. The fourth step is a conceptual layer, creating a conceptual graph for the final classification. Finally, the fifth step is the final step for predicting classes.

\begin{figure}[h]
	\centering
	 \includegraphics[width=\linewidth , keepaspectratio ]{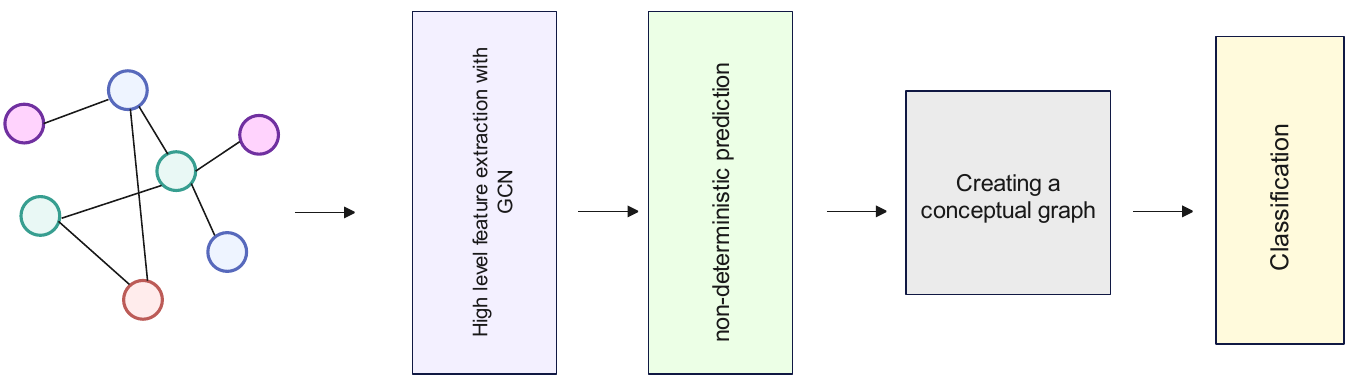}
    \caption{General schematic}
	\label{fig:fig1}
\end{figure}

Figure 4 presents the details of our proposed method. In the first step, embedding is applied to the input; in the second step, we deploy attention mechanisms (att-ls and att-rs) designed to focus the model on specific aspects or features of the input data during learning. Then, att-ls and att-rs are moved,
and the number of diagrams is entered into the encoder as input. After passing through the classifier, it enters the GCN stage. In this stage, the raw, encoded, and embedded data are fused and provided as input to the incomplete hidden layers of the initial GCN. The first GCN doesn’t engage in prediction or conclusive decision-making; instead, it focuses on
extracting high-level features. Furthermore, the output from this layer takes the form of logical (conceptual) graphs, which serve as fresh input for the second GCN. Subsequently,this layer determines the final and conclusive prediction based on these logical graphs.

\begin{figure}[h]
	\centering
	 \includegraphics[width=\linewidth  , keepaspectratio]{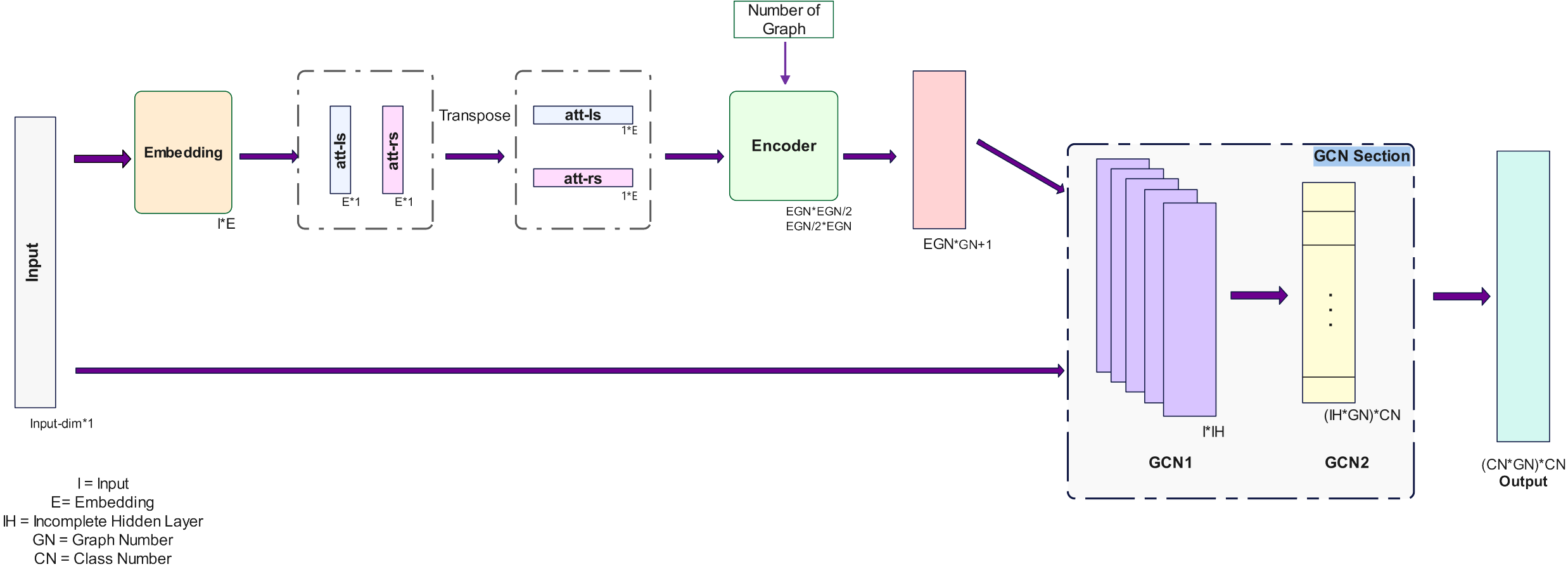}
    \caption{Details of Proposed method}
	\label{fig:fig1}
\end{figure}
\subsection{Complexity Analysis}\label{se:intro}
The computational complexity of our proposed method involves several key components. First, the adjacency matrix construction has a complexity of \( O(\mid E \mid) \), where \( \mid E \mid \) represents the number of edges in the graph. This step requires iterating through all edges to construct the matrix. Next, processing the feature matrix has a complexity of \( O(n \times m) \), where \( n \) is the number of nodes and \( m \) is the number of features, representing the cost of handling all node features. The neighborhood search, which identifies neighbors for each node, has a complexity of \( O(n \times k) \), with \( k \) being the average number of neighbors. Embedding the nodes into a new feature space has a complexity of \( O(n \times m \times d) \), where \( d \) represents the embedding dimension.

The transpose of the matrix, required in subsequent operations, has a complexity of \( O(\mid E \mid) \). Extracting attributes from the neighborhood data has a complexity of \( O(n \times k \times d) \), scaling with the node count, neighborhood size, and feature dimension. Encoding the input features results in a complexity of \( O(z^2) \), where \( z \) is the size of the final encoded representation.

The graph convolutional network (GCN) layers have a complexity of \( O(L \times \mid E \mid \times d) \), where \( L \) is the number of layers, \( \mid E \mid \) is the number of edges, and \( d \) is the feature dimension. The first GCN layer, responsible for high-level feature extraction, has a complexity of \( O(L_1 \times \mid E \mid \times d) \), while the second GCN layer, which performs the final prediction, has a similar complexity of \( O(L_2 \times \mid E \mid \times d) \), both scaling with the number of layers and the graph size.

\section{Experimental Results}\label{se:intro}
This section evaluates and comprehensively reviews our proposed method by performing various tests. All tests were performed using a system with Windows 10, a Core i7 processor, 16GB of RAM, and an NVIDIA GPU.\\

\subsection{Experimental Settings}\label{se:intro}
We evaluate the performance of our proposed model to perform the fully supervised node classification task using three benchmark datasets, the details of which are shown in Table 2. Cora, Citeseer, and PubMed are three real-world datasets commonly used for node classification and introduced by \citep{sen2008collective}.

\begin{table}[H] 
	\caption{Benchmark Graph Datasets}
	\centering
	\begin{tabular}{|l|c|c|c|}
		\toprule
		Parameter & Cora & Citeseer & PubMed \\
		\midrule
		Nodes          & 2708  & 3327     & 19,717  \\
		\hline
		Features       & 1433  & 3703     & 500     \\
		\hline
		Class          & 7     & 6        & 3       \\
		\hline
		Edges          & 5429  & 4732     & 44,338  \\
		\hline
		Content Words  & 3,880,564 & 12,274,336 & 12,274,336 \\
		\bottomrule
	\end{tabular}
	\label{tab:parameters}
\end{table}

\textbf{{\large Baseline Methods:\\}}
To demonstrate the better performance of our model, we compare it with basic methods in graph neural networks (GNN), including GCN, GAT, APPNP, GraphSAGE, and GCNII.\\
• GCN: It uses a layered propagation principle rooted in approximating the initial spectral complexity on graphs. Considering the characteristics of nodes and their neighboring connections enables the potential to obtain representations of nodes through learning \citep{kipf2016semi}.\\
• GAT: This method assigns different weights to neighboring nodes depending on their importance and is generally used as the baseline of GNN \citep{velickovic2017graph}.\\
• APPNP: This model is customized based on the relationship between GCN and PageRank and also based on the PageRank algorithm. It is designed to work with graph data and collect relational information between nodes. Its purpose is to predict labels for nodes in a graph \citep{gasteiger2018predict}.\\
• GraphSAGE: proposed for inductive node embedding that simultaneously learns the topological structure of each node’s neighborhood and the distribution of node characteristics in the neighborhood. In addition to the characteristics of nodes, it also uses the structural characteristics of graphs \citep{hamilton2017inductive}.\\
• GCNII: It is a deep GCN model capable of providing a K-order polynomial filter with arbitrary coefficients that avoid over-smoothing \citep{chen2020simple}.

\begin{table}[h]
	\caption{Summary of node classification (\%). Results for GCN, GAT, GraphSAGE are taken from \citep{maurya2022simplifying}. For APPNP and GCNII results are taken from \citep{wang2021tree}}
	\centering
	\begin{tabular}{|l|c|c|c|}
		\toprule
		Method & Cora & Citeseer & PubMed \\
		\midrule
		GCN \citep{kipf2016semi}      & 87.28 ± 1.26\%  & 76.68 ± 1.64\% & 87.38 ± 0.66\% \\
		\hline
		GAT \citep{velickovic2017graph} & 82.68 ± 1.80\% & 75.46 ± 1.72\% & 84.68 ± 0.44\% \\
		\hline
		GraphSAGE \citep{hamilton2017inductive} & 86.90 ± 1.04\% & 76.04 ± 1.30\% & 88.45 ± 0.50\% \\
		\hline
		APPNP \citep{gasteiger2018predict} & 86.76 ± 1.74\% & 77.08 ± 1.56\% & 88.45 ± 0.42\% \\
		\hline
		GCNII \citep{chen2020simple} & 88.27 ± 1.31\% & 77.06 ± 1.67\% & 90.26 ± 0.41\% \\
		\hline
		Proposed   & 96\%    & 93\%    & 95\%   \\
		\bottomrule
	\end{tabular}
	\label{tab:mytable}
\end{table}

\textbf{{\large Parameter Settings:\\}}
In this section, we briefly explain each of the parameters used. More details of the hyperparameters for all three datasets are listed in Table 4.\\
• Momentum: In deep learning, it is a parameter used to improve the training of neural networks, especially when using gradient-based optimization algorithms. Momentum helps speed up the convergence of the training process and helps improve the performance of the Loss Function. It should be noted that learning rate and momentum have almost a complementary relationship.\\
• Negative slope: It is used to prevent the decrease of the gradient in deep networks, and if the input is negative, a small non-zero coefficient is applied to it until there is a downward slope. In this way, the sudden decrease of the gradient Prevents the length of the operation.\\
• Weight-decay: The learning process of deep models gradually reduces the learning rate. This reduction of the learning rate continues until it reaches zero. This method is used to reduce the side effects associated with large weights and control the degree of generalizability of the model and is obtained from the Eqs (1):

\begin{equation}
\text{Weight-decay} = \frac{\text{Learning Rate}}{\text{Epoch}}
\label{eq:weight-decay}
\end{equation}

\noindent • Gamma coefficient (Y) or discount rate: It is a rate that controls how much a model benefits from new data changes at each stage of learning. This parameter determines how fast or slow the model is directed towards the optimal solution. We used the gamma coefficient in all three datasets.\\
• Epoch: It means a stage or period in the training of a deep model, which passes through all its training data in each period of the model.\\
• Batch Size: a meta-parameter in deep learning that determines the number of training samples used per iteration (or batch) during the training process. Due to the increase in the speed and quality of the model, it is included in the agenda of our paper.

\begin{table}[h]
	\caption{The hyper-parameters for the proposed method in three datasets}
	\centering
	\begin{tabular}{|l|c|c|c|}
		\toprule
		\textbf{Parameters} & \textbf{Cora} & \textbf{Citeseer} & \textbf{PubMed} \\
		\midrule
		Weight\_decay           & 0.0004 & 0.00035 & 0.00029 \\
		\hline
		Ratio\_node             & 0.33   & 0.56    & 0.75    \\
		\hline
		dropout                 & 0.2    & 0.4     & 0.6     \\
		\hline
		Activation Functions    & ['relu'] & ['relu'] & ['relu'] \\
		\hline
		sigma                   & 2.0    & 4.0     & 6.0     \\
		\hline
		Input\_Dim              & 1433   & 3703    & 500     \\
		\hline
		hidden layer sizes      & 16     & 32      & 64      \\
		\hline
		layer activation functions & ['LeakyReLU'] & ['LeakyReLU'] & ['LeakyReLU'] \\
		\hline
		negative\_slope         & 0.2    & 0.2     & 0.2     \\
		\hline
		Graph\_size             & 40     & 100     & 150     \\
		\hline
		Learning Rate           & 0.1    & 0.1     & 0.1     \\
		\hline
		Batch Size              & 20     & 40      & 80      \\
		\hline
		Epoch                   & 230    & 260     & 300     \\
		\bottomrule
	\end{tabular}
	\label{tab:parameters}
\end{table}

\subsection{Node Classification Results}\label{se:intro}

Table 3 shows each model’s average node classification accuracy on three selected datasets.The results show that our method increases accuracy compared to methods such as GCNII,which have been presented recently and show better performance.

\subsection{Evaluation Criteria}\label{se:intro}
Accuracy: It refers to a measure used to evaluate the performance of a model in correctly predicting the class or label of a graph. It measures the proportion of correctly classified graphs from the total number of graphs in the evaluation dataset. Accuracy is calculated by dividing the number of correctly classified graphs by the total number of graphs in the
data set. This shows how accurately the model can classify graphs. Which is calculated using Eqs (2):
\begin{equation}
\text{Accuracy} = \frac{TP + TN}{TP + TN + FP + FN}
\label{eq:accuracy}
\end{equation}
\subsection{Discussion}\label{se:intro}
Figures 5-7 show graphical representations of precision and loss curves for the Cora, Citeseer, and PubMed datasets. Graphical representations in TSNE embedding format show categories of different classes (seven, six, and three classes for Cora, Citeseer, and PubMed,respectively). Accuracy and loss curves have been drawn for the training and validation
stages, which show the x-axis of accuracy and the y-axis of loss. We obtain 96\% accuracy with a loss of 0.008 for CORA, 93\% accuracy with a loss of 0.005 for Citeseer, and 95\% accuracy with a loss of 0.0009 for PubMed.

\begin{figure}[h]
    \centering
    \renewcommand\thesubfigure{\arabic{figure}.\arabic{subfigure}} 
    \begin{subfigure}[t]{0.3\textwidth}
        \centering
        \includegraphics[width=1\linewidth , keepaspectratio ]{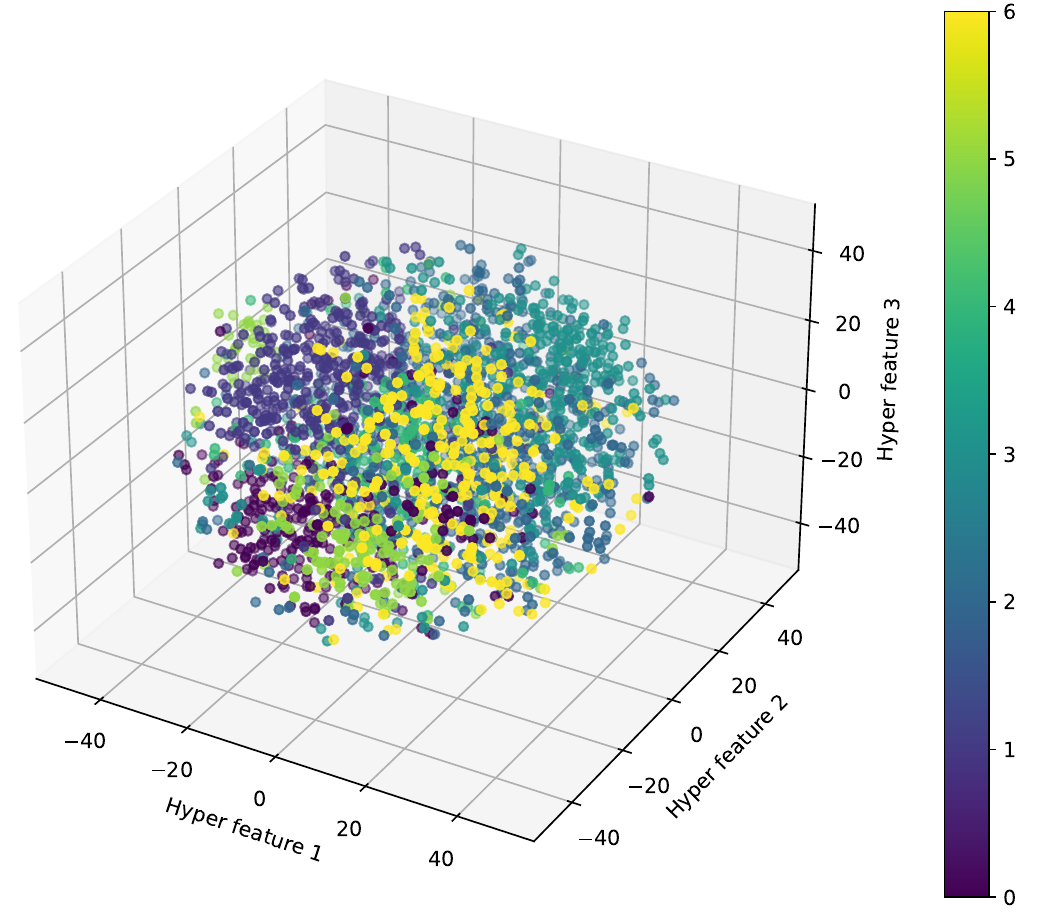}
        \caption{\small Cora classes}
        \label{fig:image1}
    \end{subfigure}\hfill
    \begin{subfigure}[t]{0.3\textwidth}
        \centering
        \includegraphics[width=1\linewidth , keepaspectratio ]{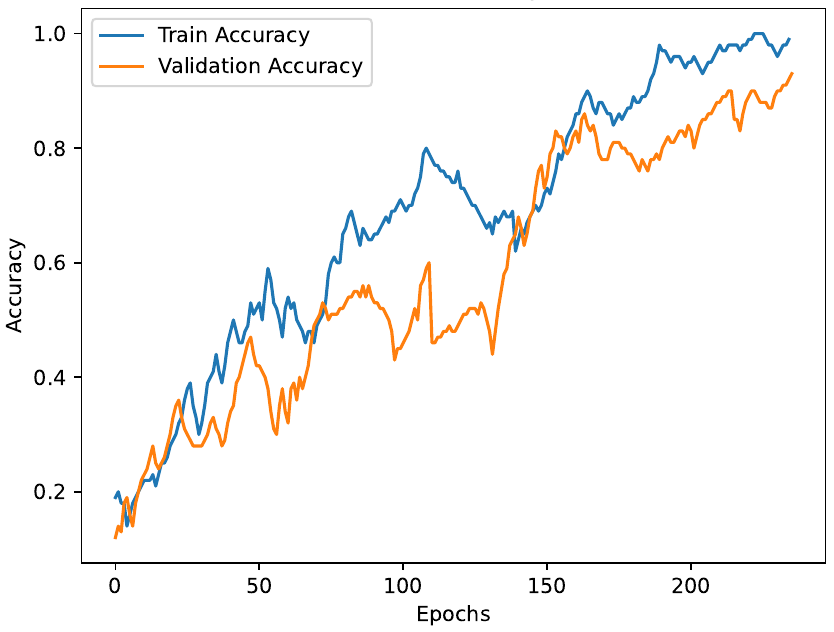}
        \caption{\small Accuracy(Cora dataset)}
        \label{fig:image2}
    \end{subfigure}\hfill
    \begin{subfigure}[t]{0.3\textwidth}
        \centering
        \includegraphics[width=1.1\linewidth , keepaspectratio ]{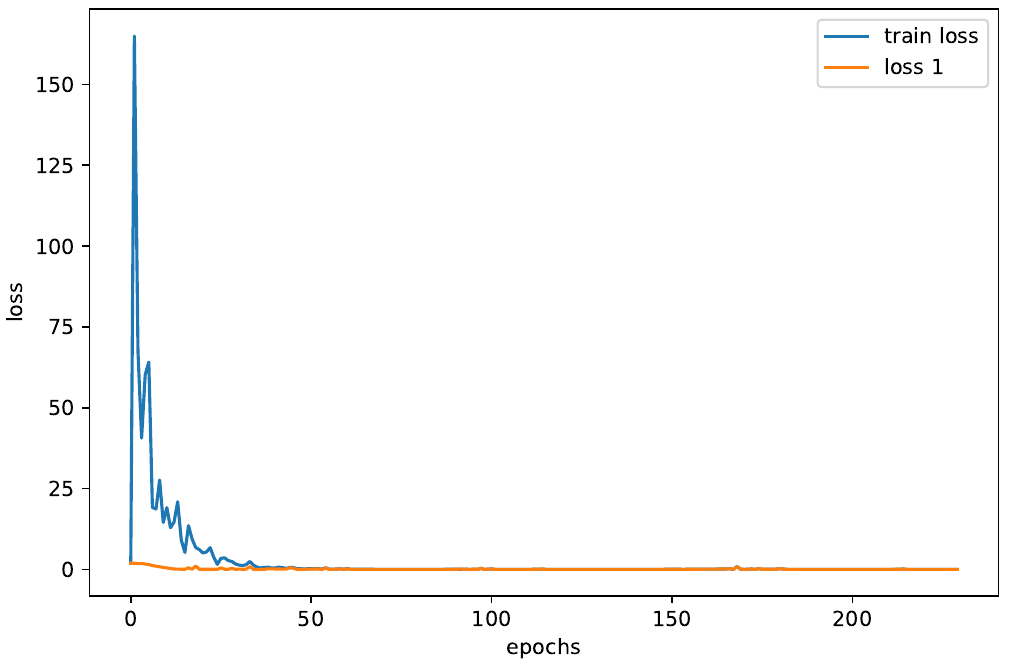}
        \caption{\small Loss(Cora dataset)}
        \label{fig:image3}
    \end{subfigure}
    \caption{Graphical representation for Cora dataset}
    \label{fig:overall1}
\end{figure}
\begin{figure}[h]
    \centering
    \renewcommand\thesubfigure{\arabic{figure}.\arabic{subfigure}}
    \begin{subfigure}[t]{0.3\textwidth}
        \centering
        \includegraphics[width=1\textwidth , keepaspectratio ]{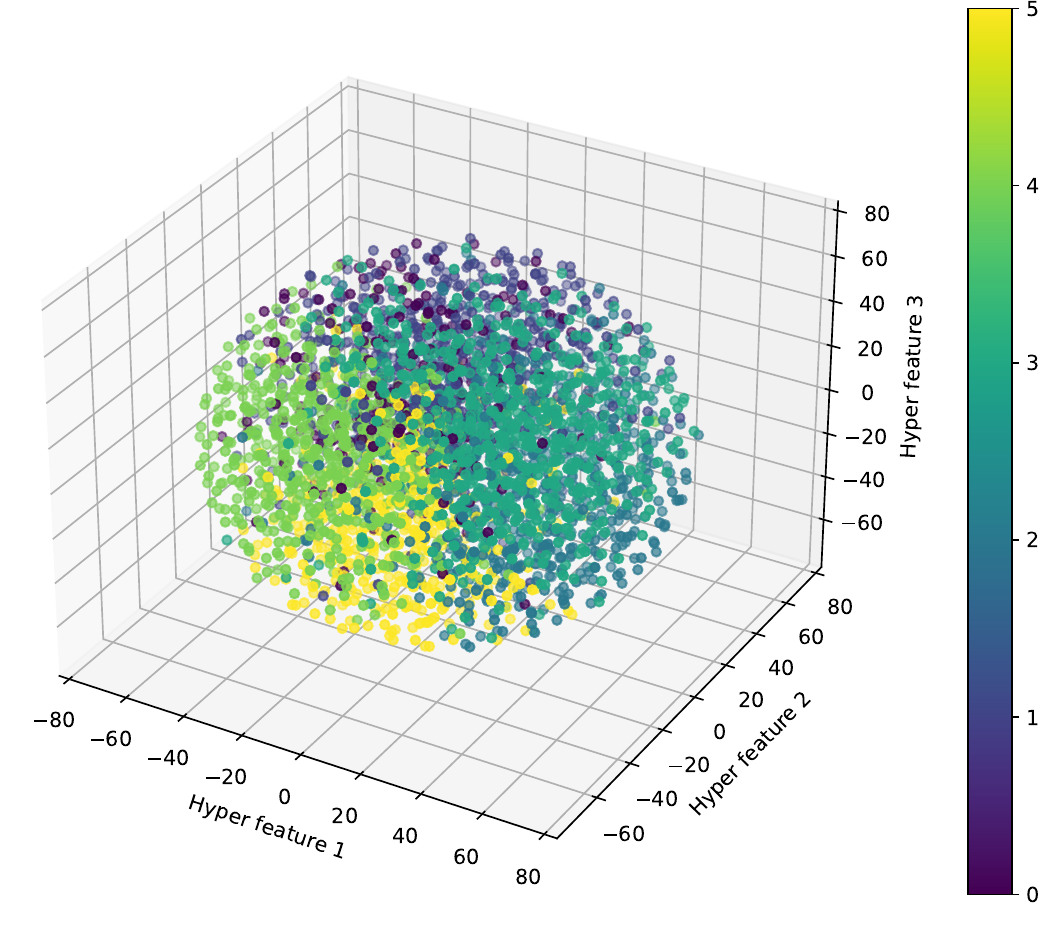}
        \caption{\small Citeseer classes}
        \label{fig:image4}
    \end{subfigure}
    \begin{subfigure}[t]{0.3\textwidth}
        \centering
        \includegraphics[width=1\textwidth ]{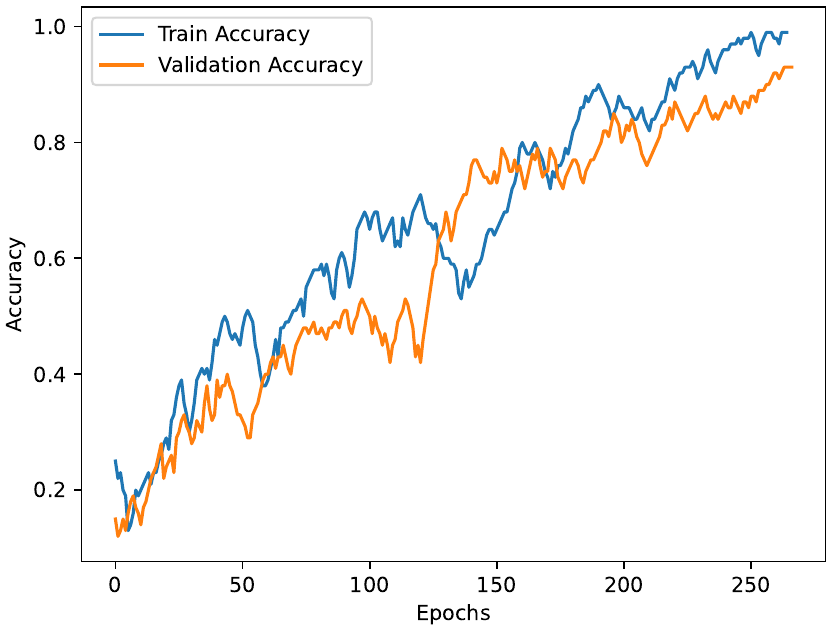}
        \caption{Accuracy (Citeseer dataset)}
        \label{fig:image5}
    \end{subfigure}
    \begin{subfigure}[t]{0.3\textwidth}
        \centering
        \includegraphics[width=1.1\textwidth  , keepaspectratio]{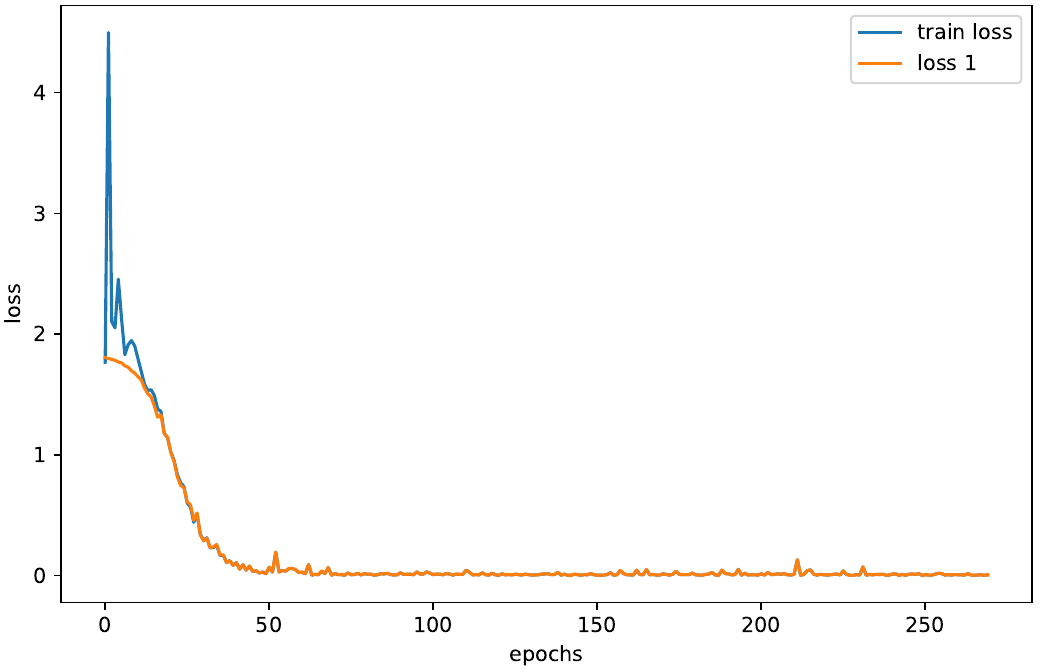}
        \caption{\small Loss(Citeseer dataset)}
        \label{fig:image6}
    \end{subfigure}
    \caption{Graphical representation for Citeseer dataset}
    \label{fig:overall2}
\end{figure}


\begin{figure}[h]
    \centering
    \renewcommand\thesubfigure{\arabic{figure}.\arabic{subfigure}}
    \begin{subfigure}[t]{0.3\textwidth}
        \centering
        \includegraphics[width=1\textwidth , keepaspectratio ]{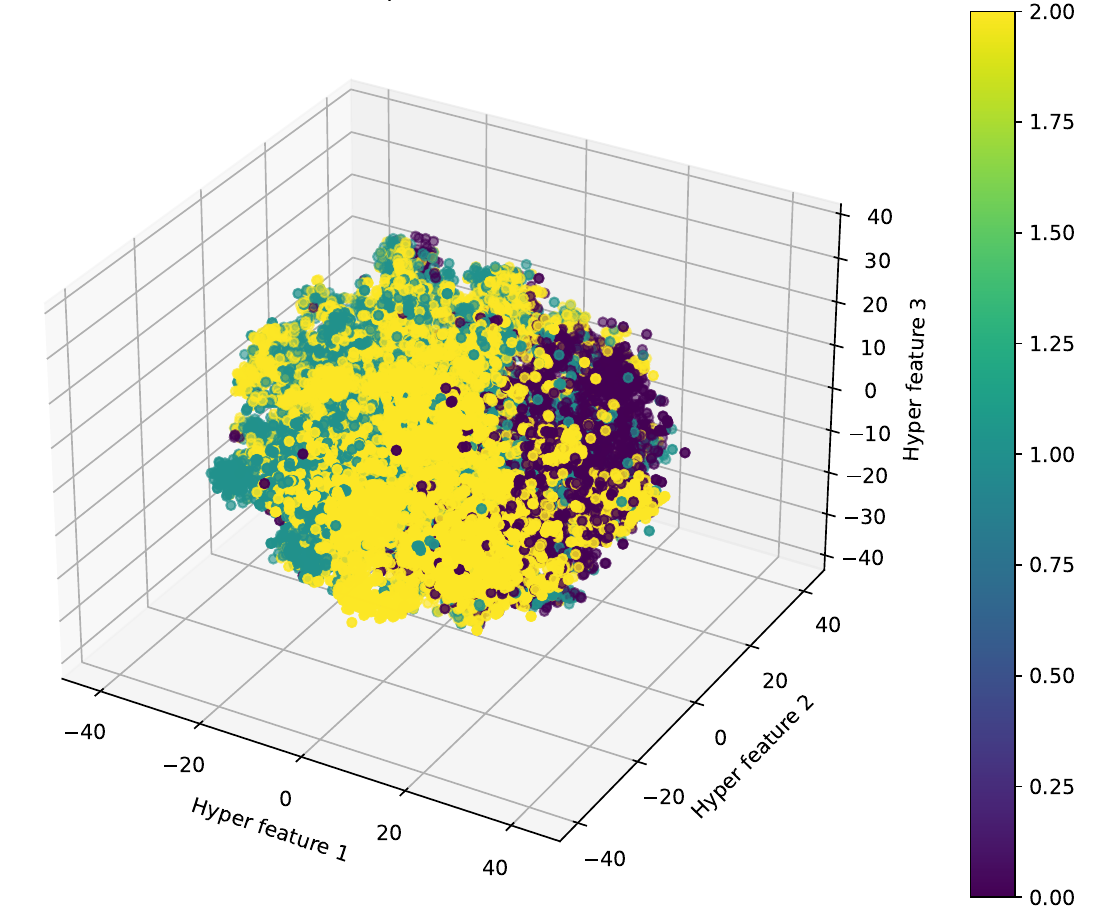}
        \caption{\small PubMed classes}
        \label{fig:image7}
    \end{subfigure}
    \begin{subfigure}[t]{0.3\textwidth}
        \centering
        \includegraphics[width=1\textwidth , keepaspectratio ]{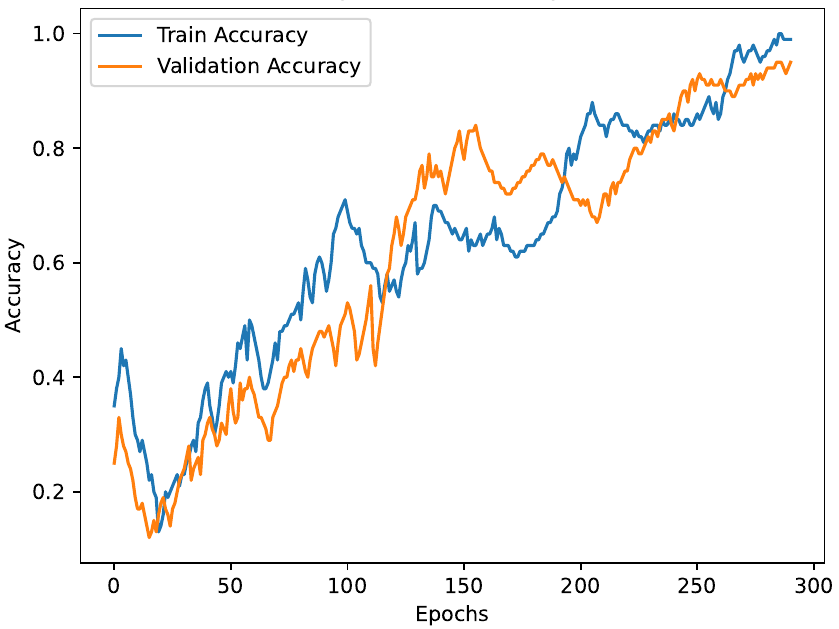}
        \caption{Accuracy (PubMed dataset)}
        \label{fig:image8}
    \end{subfigure}
    \begin{subfigure}[t]{0.3\textwidth}
        \centering
        \includegraphics[width=1\textwidth , keepaspectratio ]{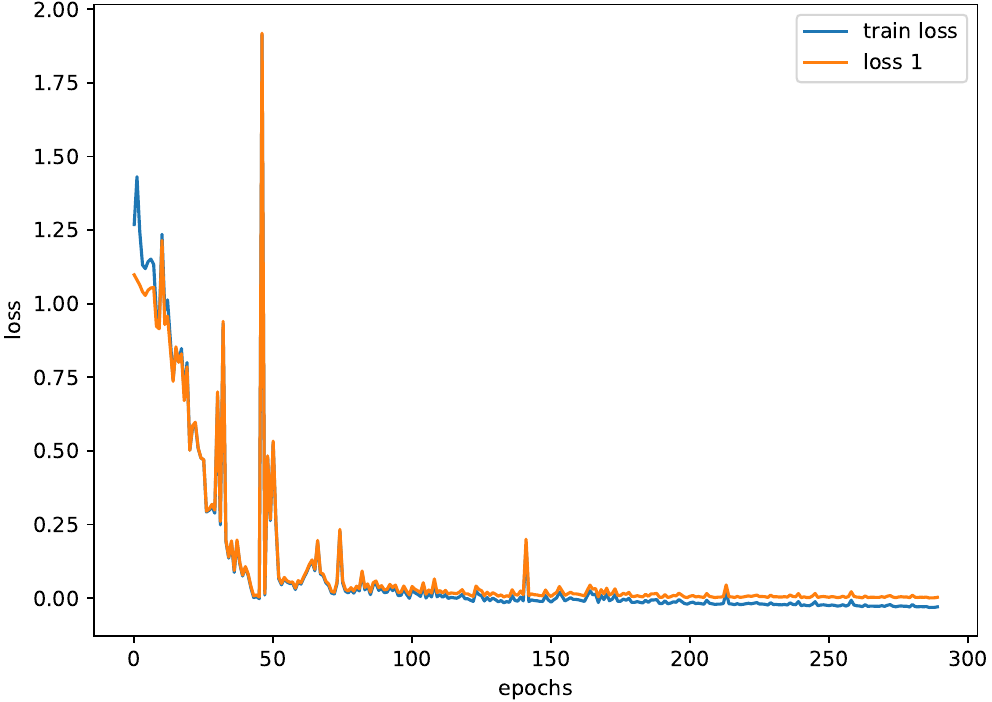}
        \caption{Loss(PubMed dataset )}
        \label{fig:image9}
    \end{subfigure}
    \caption{Graphical representation for PubMed dataset}
    \label{fig:overall3}
\end{figure}


\noindent Figure 8 illustrates the class distribution in the Cora, CiteSeer, and PubMed datasets. Concerning class distribution, the Cora dataset exhibits a lower number of samples but a higher number of classes when compared to the other two datasets. In contrast, the PubMed dataset contains more samples but fewer classes.

\begin{figure}[h]
    \centering
    \includegraphics[width=0.8\linewidth , keepaspectratio ]{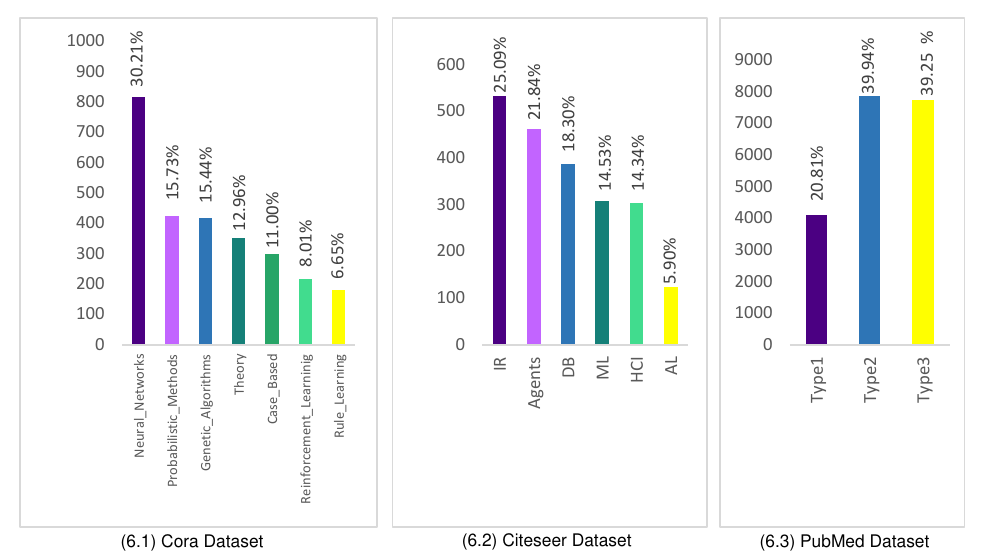}
    \caption{Datasets Distribution}
    \label{fig:example}
\end{figure}

\section{Conclusion}
In this paper, we introduced a novel approach for node classification. The proposed model comprises multiple stages. The initial stage involves utilizing GCN to derive high-level features from embedded and encoded inputs. The subsequent step employs Using incomplete (non-deterministic) models for non-deterministic forecasting. Creating a secondary conceptual graph is based on non-deterministic prediction and, finally, classification and prediction of classes. Experimental results demonstrate that our proposed method outperforms other methods by at least 5\% on benchmark datasets. Thus, in addressing the research inquiries, it can be asserted that incomplete models function as Logical classifications, and their output (logical graphs) serves as input for the ensuing step. Logical graphs enable preliminary estimations before final classification, leading to a substantial enhancement in prediction results.

\section{Future Work}

In future endeavors, the proposed approach can be applied to extract protein subpatterns in the second stage. Additionally, in social networks, it can be employed to extract diverse subgroups based on events, extract subsystems and modules of software systems, and deduce utility modules for practical applications.

\bibliographystyle{unsrtnat}
\bibliography{references}  

\begin{thebibliography}{35}
\providecommand{\natexlab}[1]{#1}
\providecommand{\url}[1]{\texttt{#1}}
\expandafter\ifx\csname urlstyle\endcsname\relax
  \providecommand{\doi}[1]{doi: #1}\else
  \providecommand{\doi}{doi: \begingroup \urlstyle{rm}\Url}\fi

\bibitem[Goyal and Ferrara(2018)]{goyal2018graph}
P.~Goyal and E.~Ferrara.
\newblock Graph embedding techniques, applications, and performance: A survey.
\newblock \emph{Knowledge-Based Systems}, 151:\penalty0 78--94, 2018.

\bibitem[Waikhom and Patgiri(2021)]{waikhom2021graph}
L.~Waikhom and R.~Patgiri.
\newblock Graph neural networks: Methods, applications, and opportunities.
\newblock \emph{arXiv preprint arXiv:2108.10733}, 2021.

\bibitem[Jiang et~al.(2020)Jiang, Zhu, Li, and Ji]{jiang2020co}
X.~Jiang, R.~Zhu, S.~Li, and P.~Ji.
\newblock Co-embedding of nodes and edges with graph neural networks.
\newblock \emph{IEEE Transactions on Pattern Analysis and Machine
  Intelligence}, 2020.

\bibitem[Pan et~al.(2019)Pan, Hu, Fung, Long, Jiang, and
  Zhang]{pan2019learning}
S.~Pan, R.~Hu, S.~Fung, G.~Long, J.~Jiang, and C.~Zhang.
\newblock Learning graph embedding with adversarial training methods.
\newblock \emph{IEEE Transactions on Cybernetics}, 50:\penalty0 2475--2487,
  2019.

\bibitem[Li et~al.(2019)Li, Rong, Cheng, Meng, Huang, and Huang]{li2019semi}
J.~Li, Y.~Rong, H.~Cheng, H.~Meng, W.~Huang, and J.~Huang.
\newblock Semi-supervised graph classification: A hierarchical graph
  perspective.
\newblock In \emph{The World Wide Web Conference}, pages 972--982, 2019.

\bibitem[Ramanath et~al.(2018)Ramanath, Inan, Polatkan, Hu, Guo, Ozcaglar, Wu,
  Krishnaram, and Cem]{ramanath2018towards}
R.~Ramanath, H.~Inan, G.~Polatkan, B.~Hu, Q.~Guo, C.~Ozcaglar, X.~Wu,
  K.~Krishnaram, and S.~Cem.
\newblock Towards deep and representation learning for talent search at
  linkedin.
\newblock In \emph{Proceedings of the 27th ACM International Conference on
  Information and Knowledge Management}, pages 2253--2261, 2018.

\bibitem[Cai et~al.(2018)Cai, Zheng, and Chang]{cai2018comprehensive}
H.~Cai, V.~W. Zheng, and K.~C. Chang.
\newblock A comprehensive survey of graph embedding: Problems, techniques, and
  applications.
\newblock \emph{IEEE Transactions on Knowledge and Data Engineering},
  30:\penalty0 1616--1637, 2018.

\bibitem[Wang et~al.(2022)Wang, Pan, Yu, Hu, Long, and Zhang]{wang2022deep}
C.~Wang, S.~Pan, P.~C. Yu, R.~Hu, G.~Long, and C.~Zhang.
\newblock Deep neighbor-aware embedding for node clustering in attributed
  graphs.
\newblock \emph{Pattern Recognition}, 122:\penalty0 108230, 2022.

\bibitem[Xu(2021)]{xu2021understanding}
M.~Xu.
\newblock Understanding graph embedding methods and their applications.
\newblock \emph{SIAM Review}, 63:\penalty0 825--853, 2021.

\bibitem[Xiao et~al.(2022)Xiao, Wang, Dai, and Guo]{xiao2022graph}
S.~Xiao, S.~Wang, Y.~Dai, and W.~Guo.
\newblock Graph neural networks in node classification: Survey and evaluation.
\newblock \emph{Machine Vision and Applications}, 33\penalty0 (4), 2022.

\bibitem[Zeng et~al.(2019)Zeng, Zhou, Srivastava, Kannan, and
  Prasanna]{zeng2019accurate}
H.~Zeng, H.~Zhou, A.~Srivastava, R.~Kannan, and V.~Prasanna.
\newblock Accurate, efficient and scalable graph embedding.
\newblock In \emph{2019 IEEE International Parallel and Distributed Processing
  Symposium (IPDPS)}, pages 462--471, 2019.

\bibitem[Yu et~al.(2021)Yu, Yang, Zhang, and Wu]{yu2021knowledge}
D.~Yu, Y.~Yang, R.~Zhang, and Y.~Wu.
\newblock Knowledge embedding based graph convolutional network.
\newblock In \emph{Proceedings of the Web Conference 2021}, pages 1619--1628,
  2021.

\bibitem[Maurya et~al.(2022)Maurya, Liu, and Murata]{maurya2022simplifying}
S.~K. Maurya, X.~Liu, and T.~Murata.
\newblock Simplifying approach to node classification in graph neural networks.
\newblock \emph{Journal of Computational Science}, 62:\penalty0 101695, 2022.

\bibitem[Hu et~al.(2021)Hu, You, Wang, Wang, Zhou, and Gao]{hu2021graph}
Y.~Hu, H.~You, Z.~Wang, Z.~Wang, E.~Zhou, and Y.~Gao.
\newblock Graph-mlp: Node classification without message passing in graph.
\newblock \emph{arXiv preprint arXiv:2106.04051}, 2021.

\bibitem[Wu et~al.(2021)Wu, Song, Huang, Ye, Xie, and Jin]{wu2021enhancing}
Y.~Wu, Y.~Song, H.~Huang, F.~Ye, X.~Xie, and H.~Jin.
\newblock Enhancing graph neural networks via auxiliary training for
  semi-supervised node classification.
\newblock \emph{Knowledge-Based Systems}, 220:\penalty0 106884, 2021.

\bibitem[Li et~al.(2020)Li, Feng, Gao, and Qiu]{li2020hierarchical}
K.~Li, Y.~Feng, Y.~Gao, and J.~Qiu.
\newblock Hierarchical graph attention networks for semi-supervised node
  classification.
\newblock \emph{Applied Intelligence}, 50:\penalty0 3441--3451, 2020.

\bibitem[Huang et~al.(2022)Huang, Tang, and Chen]{huang2022graph}
Z.~Huang, Y.~Tang, and Y.~Chen.
\newblock A graph neural network-based node classification model on
  class-imbalanced graph data.
\newblock \emph{Knowledge-Based Systems}, 244:\penalty0 108538, 2022.

\bibitem[Wang and Derr(2021)]{wang2021tree}
Y.~Wang and T.~Derr.
\newblock Tree decomposed graph neural network.
\newblock In \emph{Proceedings of the 30th ACM International Conference on
  Information and Knowledge Management}, pages 2040--2049, 2021.

\bibitem[Tang et~al.(2024)Tang, Liu, Jiang, Chen, and Dong]{tang2024hypergraph}
B.~Tang, Z.~Liu, K.~Jiang, S.~Chen, and X.~Dong.
\newblock Hypergraph node classification with graph neural networks.
\newblock \emph{arXiv preprint arXiv:2402.05569}, 2024.

\bibitem[Wang and Yang(2024)]{wang2024low}
Y.~Wang and Y.~Yang.
\newblock Low-rank graph contrastive learning for node classification.
\newblock \emph{arXiv preprint arXiv:2402.09600}, 2024.

\bibitem[Sejan et~al.(2023)Sejan, Rahman, Aziz, Baik, You, and
  Song]{sejan2023graph}
M.~A.~S. Sejan, M.~H. Rahman, M.~A. Aziz, J.~Baik, Y.~You, and H.~Song.
\newblock Graph convolutional network design for node classification accuracy
  improvement.
\newblock \emph{Mathematics}, 11:\penalty0 3680, 2023.

\bibitem[Bhattacharya et~al.(2023)Bhattacharya, Nagwani, and
  Tripathi]{bhattacharya2023communitygcn}
R.~Bhattacharya, N.~K. Nagwani, and S.~Tripathi.
\newblock Communitygcn: Community detection using node classification with
  graph convolution network.
\newblock \emph{Data Technologies and Applications}, 57:\penalty0 580--604,
  2023.

\bibitem[de~Lara and Pineau(2018)]{de2018simple}
N.~de~Lara and E.~Pineau.
\newblock A simple baseline algorithm for graph classification.
\newblock \emph{arXiv preprint arXiv:1810.09155}, 2018.

\bibitem[Lee et~al.(2018)Lee, Rossi, and Kong]{lee2018graph}
J.~B. Lee, R.~Rossi, and X.~Kong.
\newblock Graph classification using structural attention.
\newblock In \emph{Proceedings of the 24th ACM SIGKDD International Conference
  on Knowledge Discovery and Data Mining}, pages 1666--1674, 2018.

\bibitem[Wang et~al.(2021)Wang, Chen, and Chen]{wang2021egat}
Z.~Wang, J.~Chen, and H.~Chen.
\newblock Egat: Edge-featured graph attention network.
\newblock In \emph{Artificial Neural Networks and Machine Learning--ICANN
  2021}, pages 314--324, 2021.

\bibitem[Xie et~al.(2020)Xie, Yao, Gong, Chen, and Qin]{xie2020graph}
Y.~Xie, C.~Yao, M.~Gong, C.~Chen, and A.~K. Qin.
\newblock Graph convolutional networks with multi-level coarsening for graph
  classification.
\newblock \emph{Knowledge-Based Systems}, 194:\penalty0 105578, 2020.

\bibitem[Bielak et~al.(2022)Bielak, Kajdanowicz, and
  Chawla]{bielak2022attre2vec}
P.~Bielak, T.~Kajdanowicz, and N.~V. Chawla.
\newblock Attre2vec: Unsupervised attributed edge representation learning.
\newblock \emph{Information Sciences}, 592:\penalty0 82--96, 2022.

\bibitem[Zhong and Huang(2023)]{zhong2023dynamic}
Y.~Zhong and C.~Huang.
\newblock A dynamic graph representation learning based on temporal graph
  transformer.
\newblock \emph{Alexandria Engineering Journal}, 63:\penalty0 359--369, 2023.

\bibitem[Chen et~al.(2023)Chen, Xiong, Zheng, Zhang, Zhang, Jia, and
  Liu]{chen2023egc2}
J.~Chen, H.~Xiong, H.~Zheng, D.~Zhang, J.~Zhang, M.~Jia, and Y.~Liu.
\newblock Egc2: Enhanced graph classification with easy graph compression.
\newblock \emph{Information Sciences}, 629:\penalty0 376--397, 2023.

\bibitem[Sen et~al.(2008)Sen, Namata, Bilgic, Getoor, Galligher, and
  Eliassi-Rad]{sen2008collective}
P.~Sen, G.~Namata, M.~Bilgic, L.~Getoor, B.~Galligher, and T.~Eliassi-Rad.
\newblock Collective classification in network data.
\newblock \emph{AI Magazine}, 29:\penalty0 93--93, 2008.

\bibitem[Kipf and Welling(2016)]{kipf2016semi}
T.~N. Kipf and M.~Welling.
\newblock Semi-supervised classification with graph convolutional networks.
\newblock \emph{arXiv preprint arXiv:1609.02907}, 2016.

\bibitem[Velickovic et~al.(2017)Velickovic, Cucurull, Casanova, Romero, Lio,
  and Bengio]{velickovic2017graph}
P.~Velickovic, G.~Cucurull, A.~Casanova, A.~Romero, P.~Lio, and Y.~Bengio.
\newblock Graph attention networks.
\newblock \emph{stat}, 1050:\penalty0 10--48550, 2017.

\bibitem[Gasteiger et~al.(2018)Gasteiger, Bojchevski, and
  G{\"u}nnemann]{gasteiger2018predict}
J.~Gasteiger, A.~Bojchevski, and S.~G{\"u}nnemann.
\newblock Predict then propagate: Graph neural networks meet personalized
  pagerank.
\newblock \emph{arXiv preprint arXiv:1810.05997}, 2018.

\bibitem[Hamilton et~al.(2017)Hamilton, Ying, and
  Leskovec]{hamilton2017inductive}
W.~Hamilton, Z.~Ying, and J.~Leskovec.
\newblock Inductive representation learning on large graphs.
\newblock \emph{Advances in Neural Information Processing Systems}, 30, 2017.

\bibitem[Chen et~al.(2020)Chen, Wei, Huang, Ding, and Li]{chen2020simple}
M.~Chen, Z.~Wei, Z.~Huang, B.~Ding, and Y.~Li.
\newblock Simple and deep graph convolutional networks.
\newblock In \emph{International Conference on Machine Learning}, pages
  1725--1735, 2020.

\end{thebibliography}






\end{document}